\documentclass[]{interact}

\usepackage{epstopdf}
\usepackage[backend=biber,style=ieee]{biblatex}
\usepackage{algorithm}
\usepackage{graphicx}
\usepackage{multirow}
\usepackage{url}
\usepackage{xurl}
\usepackage{caption}
\usepackage{lscape} 
\usepackage{subfig}

\theoremstyle{plain}

\theoremstyle{definition}

\theoremstyle{remark}

\begin{document}

\articletype{CHAPTER PREPRINT}

\title{Fine-tuning Timeseries Predictors Using Reinforcement Learning}

\author{
\name{Hugo Cazaux\textsuperscript{a,}\textsuperscript{b} , Ralph Rudd\textsuperscript{a}, Hlynur Stefánsson\textsuperscript{a}, Sverrir Ólafsson\textsuperscript{a} and Eyjólfur Ingi Ásgeirsson\textsuperscript{a}}
\affil{\textsuperscript{a}Reykjavik University, Department of Engineering, Menntavegur 1, Reykjavik, 102, Iceland \textsuperscript{b}Corresponding author, email: hugot@ru.is}
}

\maketitle

\begin{abstract}
This chapter presents three major reinforcement learning algorithms used for fine-tuning financial forecasters. We propose a clear implementation plan for backpropagating the loss of a reinforcement learning task to a model trained using supervised learning, and compare the performance before and after the fine-tuning. We find an increase in performance after fine-tuning, and transfer learning properties to the models, indicating the benefits of fine-tuning. We also highlight the tuning process and empirical results for future implementation by practitioners.
\end{abstract}


\section{Introduction}
Timeseries predictors are generally trained using supervised learning on datasets. The standard setup divides the dataset into three segments: training, validation and testing. The model is initially fit on training data, then evaluated on the validation set to tune hyper-parameters and assess the predictive power. Finally, the test set is used by the final model to determine the accuracy on unseen data. These steps are well understood and constitute the backbone of supervised learning in timeseries prediction. 

This methodology draws a strong parallel with large language models (LLMs), which are generally transformer-based models and use supervised learning for pre-training. The pre-training is a common step to all LLMs that starts with a large amount of raw data, compressed in the network. Once the pre-training is complete, alignment aims to tune the model to create an user-friendly experience. This step incentivizes answering and asking questions to contextualize requests, teaches the LLMs how to use external tools, or censors potentially harmful information that might lie within the embeddings. The novelty of this research lies in the extension of alignment to time-series prediction model. 

Fine-tuning in large language models was initially based on human feedback. A standard setup consists of a human operator prompting a question and grading the answer. Later, practitioners aimed at removing subjectivity from the fine-tuning pipeline by instead proposing two answers to a prompt and have a human operator select the best one. These two methods fall under the umbrella of Reinforcement Learning with Human Feedback (RLHF), and although effective recent models have shown pure Reinforcement Learning (RL) approaches outperforming RLHF for a fraction of the cost. 

The central idea of this chapter is to leverage the predictive power of supervised pre-training and to use RL algorithms to align the model with diverse constraints. These constraints can be domain specific, such as risk management or operational constraints, but can also be purely mathematical, such as incentivizing bolder out-of-sample predictions. The reward function is at the center of the tuning, and will determine which direction the model is pushed towards. This approach is more adapted to time-series prediction compared to RLHF, as it completely removes the human operator and the need to reduce subjectivity in the feedback. RL is also extremely cost effective, since it removes the need of a coordinated effort of human operators giving feedback on a large number of samples.

In the context of time-series prediction, RL for fine-tuning is novel. The standard implementations of reinforcement learning in time-series prediction consist of a completely untrained agent learning a policy over a simulated environment. In the case of finance, this environment might be a portfolio or an ensemble of assets. In this chapter, we used a pre-trained model from \cite{cazaux2025nsi} that serves as a backbone for the RL implementation. The environment is set up to reflect the training data closely, with the main tuning tool available being the reward structure. The loss is back-propagated through the backbone, updating the weights according to the policy. 

The research questions investigated in this chapter are: Can we fine-tune pre-trained models to enhance time-series predictions using reinforcement learning? What state-of-the-art reinforcement algorithms work best for fine-tuning? 

This chapter is structured as follow: Section \ref{sec:background_ft} presents a literature review, section \ref{sec:data} the data used to train/test the models, Section \ref{sec:framework_ft} the framework used to fine-tune and evaluate the models, Section \ref{sec:benchmark_ft} benchmarks the models on standard reinforcement learning tasks, Section \ref{sec:results_ft} the results of the fine-tuning, Section \ref{sec:tuning_ft} the tuning of the specific hyperparameters and finally Section \ref{sec:conclusion_ft} is the conclusion to the chapter.

\section{Background}
\label{sec:background_ft}
Fine-tuning has become an emerging trend since large pre-trained model became more widely available to the public \cite{church2021emerging}. Fine-tuning is a technique that intends to specialize a pre-trained backbone model, often to increase performance on selected benchmarks \cite{tajbakhsh2016convolutional} or to benefit from previously acquired knowledge through transfer learning \cite{howard2018universal}. The democratization of open source models with available weights in natural language processing \cite{llama}, \cite{devlin2018bert} and image processing \cite{rombach2022high} enabled researchers and enthusiasts to propose their own fine-tuned version of an advanced model without the high computational cost of pre-training. Fine-tuning was leveraged to propose fine art classification \cite{cetinic2018fine}, fine-tuning large language models for better medical care \cite{xiong2023doctorglm}, biomedical tasks in different languages \cite{luo2024taiyi}, and malware detection in images \cite{vasan2020imcfn}.

As the size of models and the parameter number grow exponentially, fine-tuning the entire model for each downstream tasks was replaced with a sparser approach called parameter-efficient fine-tuning \cite{xu2023parameter}, \cite{fu2023effectiveness}. Methods such as Adapter \cite{zhang2023llama}, \cite{he2021effectiveness}, LoRA \cite{hu2021lora} and Prefix-tuning \cite{li2021prefix} propose to modify the architecture of the original model to benefit from higher order patterns learned during supervised learning while also specializing in a downstream task. Supervised fine-tuning uses labeled data after pre-training to align the model towards a downstream task. This method has grown in popularity as large language models hit the public sphere and adapted for more intuitive or safer usage \cite{gunel2020supervised}, \cite{zhou2021closer}, \cite{zhang2020revisiting}.

As the cost of computation carried over to efficient data labeling \cite{fredriksson2020data}, alternative techniques for fine-tuning were explored. Reinforcement learning, one of the major paradigms in machine learning, has become one of the prime candidate for efficient fine-tuning. Adversarial networks had previously shown promising results \cite{chen2020adversarial}, and policy learning has been employed in text-to-image \cite{fan2023dpok} and multi-modal models \cite{zhai2025fine}. Perhaps the most impressive implementation of reinforcement learning based fine-tuning comes from the DeepSeek-v3 report \cite{liu2024deepseek}, which implements group proximal policy optimization to fine-tune a pre-trained model and implement chain-of-thoughts reasoning. 

Within time-series prediction, fine-tuning has been focused on domain adaption. In a similar fashion to text and image generation, large pre-trained models are becoming available to researchers \cite{liutimer}. The models can then be fine-tuned for domain specific predictions and receive the same benefit as large language models \cite{chang2023llm4ts}, \cite{liang2024foundation}. However, these methods involve supervised fine-tuning, which in the case of time-series prediction consists of adding data form the specific domain the model needs to be fine-tuned on. As large language models have proven in the past, this method of fine-tuning can quickly become unsustainable due to the increasing cost of data labeling. In this study, we follow the way paved by LLMs by proposing reinforcement learning to tune time-series predictors.

PPO is a policy gradient method developed by John Schulman et al. in 2017 \cite{Schulman2017}. The key innovation of this algorithm over older methods such as TRPO \cite{schulman2015trust} or ACER \cite{wang2016sample} is the clip function that constrains policy updates of the agent. PPO has been used in a wide variety of applications: Atari games \cite{kaiser2019model}, track racing games \cite{holubar2020continuous}, suspension monitoring for cars \cite{han2022reinforcement}, and image captioning \cite{zhang2021image}. A number of articles have proposed innovations to the base algorithm, for instance an alternative minimization target \cite{kobayashi2021proximal}, \cite{gu2021proximal} introduced policy feedback; specifically improving early learning stages, which are recognized as a potential weak point of PPO \cite{hsu2020revisiting}. Recently proposed improvements include a shift in learning to offline policy optimization \cite{cai2020provably} and including conservatism \cite{yu2021combo}.

Multi-agent methods have gained significant attention in the field of reinforcement learning, particularly for their capability to simulate complex systems involving interactive agents. A notable early work in multi-agent systems is \cite{tan1993multi} which explored the dynamics of cooperative and competitive agents in a shared environment. Recent advancements have integrated PPO into multi-agent applications: \cite{lowe2017multi} applied multi-agent PPO to competitive and cooperative tasks, \cite{berner2019dota} successfully employed multi-agent reinforcement learning in the complex environment of the Dota 2 game. The integration of PPO into multi-agent systems has also been explored in real-world scenarios such as traffic light control \cite{liang2019deep}, and collaborative robotics \cite{matignon2012coordinated}. Innovations specific to multi-agent PPO include \cite{yu2020meta} which introduced a meta-learning approach to enhance adaptability across different tasks and agent configurations and \cite{palmer2018lenient}, which presented the concept of leniency in multi-agent learning, mitigating the non-stationary issue commonly faced in such environments.

Attention is a machine learning mechanism designed to imitate human awareness. Attention was brought to the forefront of the field with the transformer architecture, a self-attention-based architecture that enabled the recent breakthroughs in large language models \cite{vaswani2017attention}. It has since seen many implementations including in recurrent neural networks for search results customization \cite{guo2019single}, missing data imputation \cite{wu2020attention}, and in computer vision \cite{ER2016388}. In reinforcement learning, attention models have been developed within theoretical frameworks \cite{bramlage2022generalized} and diverse applications such as source code summarizing \cite{wang2020reinforcement}, dynamic graph problems \cite{gunarathna2022solving}, and road networks management \cite{liu2024jointppo}.

The novelty of the framework presented lies in the combination of staple reinforcement learning models with time-series predictors. This chapter also creates an opportunity for further applications of the framework in simulated environment encompassing diverse fields.

\section{Data}
\label{sec:data}
To contextualize the fine-tuning we detail the financial datasets used to train the backbone and to build the fine-tuning environment. We also present the MuJoCo framework, which we use to benchmark pure reinforcement learning performance between algorithms.

\subsection{Financial and ESG Data}
\label{sub:financial_data}

The financial and ESG data used in this chapter span from intraday market prices to annual sustainability ratings. Our primary sources are:

\begin{itemize}
  \item \textbf{Refinitiv} \cite{reuters}: a global leader in financial data and analytics, covering over 80\% of global market capitalization with more than 450 ESG metrics.  We extract daily price and volume data via Refinitiv Eikon, together with the three ESG pillar scores (Environmental, Social, Governance) and the combined ESG score.
  \item \textbf{Sustainalytics} \cite{sustainalytics}: provides ESG Risk Ratings for listed firms, widely used by asset managers and banks to construct sustainable portfolios.  We incorporate their flagship ESG Risk Ratings into our dataset.
  \item \textbf{SASB Standards} \cite{ifrs,soderstrom2007ifrs}: the Sustainability Accounting Standards Board identifies material sustainability issues by industry.  Since August 2022, SASB standards underline IFRS S1 and S2 disclosures.  We one-hot encode each firm’s material SASB issue set based on the 2018 publication.
\end{itemize}

Table~\ref{tab:samp_aapl_fin} shows a snippet of Apple’s daily price data from 2005-12-05 to 2005-12-13.  The full time span of the dataset is 2005-12-05 through 2024-08-07.

\begin{table}[H]
  \centering
  \begin{tabular}{lrrrrr}
    \hline
    \textbf{Date} & \textbf{Open} & \textbf{Low} & \textbf{High} & \textbf{Close} & \textbf{Volume} \\
    \hline
    2005-12-05 & 2.17 & 2.15 & 2.19 & 2.16 & 5.84e8 \\
    2005-12-06 & 2.23 & 2.21 & 2.25 & 2.23 & 8.57e8 \\
    2005-12-07 & 2.24 & 2.20 & 2.24 & 2.23 & 6.79e8 \\
    2005-12-08 & 2.21 & 2.19 & 2.23 & 2.23 & 7.90e8 \\
    2005-12-09 & 2.24 & 2.21 & 2.25 & 2.24 & 5.55e8 \\
    2005-12-12 & 2.26 & 2.25 & 2.27 & 2.26 & 5.25e8 \\
    2005-12-13 & 2.25 & 2.24 & 2.27 & 2.26 & 4.94e8 \\
    \hline
  \end{tabular}
  \caption{Sample daily financial data for AAPL}
  \label{tab:samp_aapl_fin}
\end{table}

To enrich the raw price and volume data, we compute:
\begin{itemize}
  \item \emph{Log returns}, controlling for market effects via the Fama–French 5 factors \cite{fama2015}.
  \item Technical indicators from historical prices and volumes: 
    \begin{itemize}
      \item Relative Strength Index (RSI) \cite{belafsky2002validity},
      \item Moving Average Convergence Divergence (MACD) \cite{chong2008technical},
      \item Bollinger Bands \cite{bollinger1992using}.
    \end{itemize}
\end{itemize}
The target variable is the FF5-adjusted log return, following the methodology of \cite{cazaux2024correlation}. Financial data are available at sub-daily frequency, whereas ESG scores refresh annually (Refinitiv) or “regularly” (Sustainalytics). We evaluated regression, interpolation, autoencoders and forward-fill strategies. To respect provider methodologies and avoid compounding model error, we adopt a forward-fill approach for ESG values between update dates.

\subsection{MuJoCo Benchmarking Environments}
\label{sub:mujoco_data}

Multi-Joint dynamics with Contact, commonly called MuJoCo \cite{todorov2012mujoco}, proposes several standard environments to train and benchmark models on. To evaluate pure reinforcement learning performance, we employ three standard MuJoCo tasks:
\begin{itemize}
  \item \textbf{HalfCheetah-v4},
  \item \textbf{Hopper-v4},
  \item \textbf{Humanoid-v4}.
\end{itemize}
MuJoCo provides a high-fidelity physics simulator for continuous-control benchmarks, where:
\begin{itemize}
  \item \emph{State} \(s_t\in\mathbb{R}^{d}\) consists of joint angles, velocities and (for Humanoid) contact forces.
  \item \emph{Action} \(a_t\in\mathbb{R}^{m}\) represents torque inputs to each joint.
  \item \emph{Reward} combines forward progress, control costs, and (where applicable) healthy posture and contact penalties.
\end{itemize}

\begin{table}[H]
  \centering
  \begin{tabular}{l l}
    \hline
    \textbf{Environment} & \textbf{Reward} \\
    \hline
    HalfCheetah-v4 & \(R = w_f F - w_{\mathrm{ctrl}} \, C\) \\
    Hopper-v4      & \(R = w_f F + w_h H - w_{\mathrm{ctrl}} \, C\) \\
    Humanoid-v4    & \(R = w_f F + w_h H - w_{\mathrm{ctrl}} \, C - w_{\mathrm{ctct}} \, C_{\mathrm{tct}}\) \\
    \hline
  \end{tabular}
  \caption{MuJoCo environment reward functions (forward reward \(F\), healthy reward \(H\), control cost \(C\), contact cost \(C_{\mathrm{tct}}\))} 
  \label{tab:mujoco_envs}
\end{table}

Here, \(w_f, w_h, w_{\mathrm{ctrl}}, w_{\mathrm{ctct}}\) are environment-specific weights.  We use the default observation and action spaces as defined in OpenAI Gym’s MuJoCo suite.

\section{Framework Details}
\label{sec:framework_ft}
As mentioned in \cite{engstrom2020implementation}, implementation is key in deep policy gradient algorithms. As such, the framework below is implemented using the clean-rl library \cite{huang2022cleanrl}. We evaluate three state-of-the-art algorithms for fine-tuning: Proximal Policy Optimization (PPO), Centralized Multi-Agent PPO (CMAPPO), and Group Relative Policy Optimization (GRPO). In this section, we also detail the environment used during training and the integration of the pre-trained transformer in the algorithms. 
\subsection{Proximal Policy Optimization (PPO)}
\begin{itemize}
    \item \textbf{Policy Function:} For an agent $x$, its policy at time $t$ is a probability density function denoted as $\pi_{\theta}(a_{t} | o_{t})$, where $\theta$ are the parameters of the policy, $o_{t}$ is the observation for agent $x$ at time $t$, and $a_{t}$ are the actions that can be taken. The policy is then sampled to obtain the action taken $\alpha_t \sim \pi_{\theta}(a_{t} | o_{t})$.

    \item \textbf{Objective Function:} The PPO objective function is defined as:
   \begin{equation*}
          L^{PPO}(\theta) = \mathbb{E}_t\left[\min(r_t(\theta) \hat{A}_t, \text{clip}(r_t(\theta), 1 - \epsilon, 1 + \epsilon) \hat{A}_t)\right]
   \end{equation*}
   where $r_t(\theta) = \frac{\pi_{\theta}(a_{t} | o_{t})}{\pi_{\theta_{\text{old}}}(a_{t} | o_{t})}$ is the probability ratio, $\epsilon$ an hyperparameter and $\hat{A}_t$ is an estimator of the advantage at time $t$, typically computed using Generalized Advantage Estimation (GAE).

   \item Advantage Estimation: The advantage $\hat{A}_t$ is computed as:
   \begin{equation}
          \hat{A}_t = \delta_t + (\gamma \lambda) \delta_{t+1} + \ldots + (\gamma \lambda)^{T-t+1} \delta_{T-1}
   \end{equation}

   with $\delta_t = r_{t} + \gamma V(o_{t+1}) - V(o_{t})$ and $V$ a learned state-value function.

   \item \textbf{Training Process:} The agent is trained by iteratively updating its policy parameters. This involves:
\begin{enumerate}
    \item Collecting trajectories by interacting with the environment using the current policy.
    \item Estimating the advantages using GAE.
    \item Calculating the surrogate objective function.
    \item Optimizing the surrogate objective function using gradient ascent while ensuring the updates stay within a specified clipping range to maintain policy stability.
\end{enumerate}
\end{itemize}

\subsection{Centralized Multi-Agent PPO (CMAPPO)}
\begin{itemize}
  \item \textbf{Subagent Policy \& Training:}  
    Each subagent \(x_i\) observes its local state \(o_{t,i}\), samples an action  
    \(\alpha_{t,i}\sim\pi_{\theta_i}(a_{t,i}\mid o_{t,i})\), and learns via its own reward \(R_i(o_t,a_{t,i})\) using PPO:
    \begin{enumerate}
      \item \textit{Collect trajectories:} Interact with environment to gather \(\{(o_{t,i},\alpha_{t,i},r_{t,i})\}_{t=1}^T\}\).  
      \item \textit{Advantage estimation:} Compute \(\hat A_{t,i}\) via GAE:  
        \(\hat A_{t,i}=\delta_{t,i}+(\gamma\lambda)\delta_{t+1,i}++(\gamma\lambda)^{2}\delta_{t+2,i}+\cdots\), with \(\delta_{t,i}=r_{t,i}+\gamma V(o_{t+1,i})-V(o_{t,i})\).  
      \item \textit{Surrogate objective:} 
        \[
          L_i^{\mathrm{PPO}}(\theta_i)=\mathbb{E}_t\!\big[\min(r_{t,i}\hat A_{t,i},\;\mathrm{clip}(r_{t,i},1-\epsilon,1+\epsilon)\hat A_{t,i})\big],
        \]  
        where \(r_{t,i}=\frac{\pi_{\theta_i}}{\pi_{\theta_i^\mathrm{old}}}\).  
      \item \textit{Policy update:} Perform gradient ascent on \(L_i^{\mathrm{PPO}}\), clipping updates to maintain stability.
    \end{enumerate}
  \item \textbf{Attention‐Enhanced Aggregation:}  
    Encode the global state \(e_t\) and subagent actions \(\{\alpha_{t,i}\}\) via linear layers, compute attention weights  
    \([w_{\mathrm{env}},w_1,\dots,w_n]=\mathrm{softmax}([f_{\mathrm{env}}(e_t),\,f_{\mathrm{sub}}(\{\alpha_{t,i}\})])\),  
    then aggregate:
    \[
      d_t = w_{\mathrm{env}}\,e_t \;+\;\sum_{i=1}^n w_i\,\alpha_{t,i}.
    \]
  \item \textbf{Superagent Decision:}  
    The superagent samples its final action  
    \(\alpha_t^f\sim\pi_{\theta_f}(a_t^f\mid d_t)\),  
    allowing coordinated, adaptive decisions across all agents.
\end{itemize}

\subsection{Group Relative Policy Optimization (GRPO)}
\begin{itemize}
    \item \textbf{Policy Function:} 
    As in PPO, we parameterize a stochastic policy 
    \(\pi_{\theta}(a \mid o)\) with parameters \(\theta\). 
    At each step \(t\), given observation \(o_{t}\), we sample a group of \(G\) candidate actions
    \[
      a_{t,i} \sim \pi_{\theta}(\,\cdot\,\mid o_{t}), \quad i=1,\dots,G.
    \]

    \item \textbf{Group Rewards and Relative Advantage:} 
    Each candidate action \(a_{t,i}\) is scored by a reward function \(r(a_{t,i},o_{t})\), yielding
    \[
      r_{t,i} = r(a_{t,i},o_{t}).
    \]
    We compute the group baseline (mean) and standard deviation:
    \[
      \bar r_{t} = \frac{1}{G}\sum_{i=1}^{G} r_{t,i}, 
      \quad 
      \sigma_{t} = \sqrt{\frac{1}{G}\sum_{i=1}^{G}\bigl(r_{t,i}-\bar r_{t}\bigr)^{2}} + \epsilon.
    \]
    The \emph{relative advantage} of candidate \(i\) is then:
    \[
      A_{t,i} \;=\; \frac{r_{t,i} - \bar r_{t}}{\sigma_{t}}.
    \]

    \item \textbf{Surrogate Objective:} 
    Defining the probability ratio for each candidate,
    \[
      \rho_{t,i}(\theta)
      = \frac{\pi_{\theta}(a_{t,i}\mid o_{t})}
             {\pi_{\theta_{\mathrm{old}}}(a_{t,i}\mid o_{t})},
    \]
    the GRPO loss uses the same clipped surrogate as PPO but averages over the group:
    \[
      L^{\mathrm{GRPO}}(\theta)
      = \mathbb{E}_{t}\!\Biggl[\,
          \frac{1}{G}\sum_{i=1}^{G}
            \min\!\bigl(\rho_{t,i}(\theta)\,A_{t,i},\;
                        \mathrm{clip}\bigl(\rho_{t,i}(\theta),1-\epsilon,1+\epsilon\bigr)\,A_{t,i}
                  \bigr)
      \Biggr].
    \]
    Optionally, one may add a KL‐penalty term 
    \(\beta\,D_{\mathrm{KL}}\bigl(\pi_{\theta}(\cdot\mid o_{t})\|\pi_{\mathrm{ref}}(\cdot\mid o_{t})\bigr)\) 
    to constrain policy drift.

    \item \textbf{Training Process:} 
    GRPO proceeds in iterative updates:
    \begin{enumerate}
        \item \textit{Sample Groups:} For each observation \(o_{t}\) in a batch, sample \(G\) actions \(\{a_{t,i}\}\).
        \item \textit{Evaluate Rewards:} Compute \(r_{t,i}=r(a_{t,i},o_{t})\) for \(i=1\ldots G\).
        \item \textit{Compute Advantages:} Form relative advantages \(A_{t,i}=(r_{t,i}-\bar r_{t})/\sigma_{t}\).
        \item \textit{Surrogate Update:} Optimize \(\theta\) by ascending the clipped surrogate \(L^{\mathrm{GRPO}}(\theta)\) (plus optional KL term), using minibatch gradient steps.
        \item \textit{Repeat:} Collect new groups under the updated policy and continue until convergence.
    \end{enumerate}
\end{itemize}

\subsection{Design of the Reinforcement Learning Environment}
The RL environment is designed to facilitate the fine-tuning of forecasting policies:
\begin{itemize}
    \item \textbf{State:} At time $t$, the state $s_t \in \mathbb{R}^{T \times N}$ is a matrix containing historical observations.
    \item \textbf{Agent Action:} The agent produces a forecast $a_{t,i} \in \mathbb{R}^{P \times 1}$ based on its local observation $o_{t,i}$.
    \item \textbf{Transition Dynamics:} Following the agents' actions, the true future $y_t \in \mathbb{R}^{P \times 1}$ is revealed, and the state is updated (via a sliding window mechanism).
    \item \textbf{Reward:} The reward $r_t$ is computed based on the forecast error and any additional domain-specific criteria:
    \begin{equation}
        r_t = -\ell(a_t, y_t) - \psi(a_t),
    \end{equation}
    where $\ell(\cdot)$ is an error metric (e.g., absolute or squared error) and $\psi(\cdot)$ encapsulates further constraints or penalties. 
\end{itemize}
In practice, the reward function used was $r_t =2 \times e^{(-MSE(a_t, y_t)} - 1$. This implementation constrains the reward between $[-1, 1]$, and is driven up as the MSE converges towards 0.

\subsection{Latent Representation versus Actor Network}
\label{sub:latent_vs_actor}
In practice, the probability distribution each of the algorithms sample from is a neural network. In a classic reinforcement learning approach, a new network is created to learn the latent representation between observations and actions (the action network). In the case of PPO and CMAPPO, networks are also created to learn the value function (the critic network). To fine-tune a pre-trained backbone model, we need to integrate the trained network in the framework. There are two main paradigms for fine-tuning the network:
\begin{itemize}
    \item \textbf{The backbone outputs a latent representation of the observation space.} The action network takes the latent representation as input and outputs a probability distribution over actions, which when sampled outputs the forecast. The critic network estimates the state value for advantage estimation and the gradients flow back through the action network, critic network, and the backbone, which leads to fine-tuning.
    \item \textbf{The backbone is connected to a projection layer that converts the latent representation to a forecast directly.} This is what commonly happens when the backbone is used independently as a predictor. In this paradigm, the backbone takes the place of the actor network. The critic network estimates the state value and the gradients flow back through the backbone and the critic network. 
\end{itemize}

Using a separate action network can improve the flexibility since the actor network has the opportunity to learn from the latent features. Decoupling the backbone and the action network also allows us to adjust the hyperparameters for the action network individually. An actor network is also more likely to explore and better adapt to the reward structure of the environment, performing significantly better in the reinforcement learning environment. We can also delay the fine-tuning by temporarily freezing all the backbone layers. This can be beneficial to performance as it gives the opportunity for the action and value networks to learn about the environment before inducing changes in the backbone network. This process can help avoid catastrophic forgetting during the early stages of interacting with the environment. 

By replacing the actor network with the backbone, we ensure that a new actor network will not corrupt the original predictor. This approach is simpler and more direct, as the actor network introduces new hyperparameters but directly using the backbones only involves a minor projection. With no actor network involved, there is also less risk of overfitting the reinforcement learning task, thus maintaining a good degree of generalization. However, without an intermediary network to adapt the learned features, the backbone might struggle to perform and learn in the reinforcement learning environment. This can lead to repeated poor performance which in turn can flow through the gradient and cause catastrophic forgetting. The environment also needs to be carefully designed to avoid a mismatch between the observations at each step of the training and the encoder size of the backbone. 

Both methods are compared in Table \ref{tab:latent_vs_actor} using standard PPO. The reference scores are the scores of the backbone without any fine-tuning. The latent paradigm performs significantly worse, with only a small improvement in the Financial sector and massive loss in Industrials and Technology. The Actor paradigm improves upon the reference on all datasets. As such, we implemented the actor paradigm when possible. The only latent representation used was in CMAPPO with the superagent, as the aggregation of the subagents action does not correspond to the encoder accepted size of the backbone.
\begin{table}[H]
\centering
\caption{Latent vs Actor paradigms comparison. The backbone is fine-tuned using PPO on Financial, Industrials and Technology. Reference is the base model without fine-tuning. Lower is better, in bold the best metric.}
\label{tab:latent_vs_actor}
\begin{tabular}{lrrrrrr}
\hline
Dataset &  \multicolumn{2}{c}{Latent} &  \multicolumn{2}{c}{Actor}  &  \multicolumn{2}{c}{Reference}  \\
Metric & \multicolumn{1}{c}{MSE} & \multicolumn{1}{c}{MAE} & \multicolumn{1}{c}{MSE} & \multicolumn{1}{c}{MAE} & \multicolumn{1}{c}{MSE} & \multicolumn{1}{c}{MAE} \\ \hline
Financial  &  0.202 & 0.206 & \textbf{0.200} & 0.271   & 0.203 & \textbf{0.118} \\
Industrials & 0.274 & 0.251 & \textbf{0.119} & \textbf{0.116}  & 0.128 & 0.121 \\
Technology & 0.341 & 0.264 & \textbf{0.126} & \textbf{0.119}  & 0.131 & 0.119\\
\hline
\end{tabular}
\end{table}

\section{Benchmarking}
\label{sec:benchmark_ft}
Three MuJoCo environments were selected as experimental settings. The three environments are: Hopper-v4, Half-Cheetah-v4 and Humanoid-v4. In this experiment, we use standard 64 hidden dimensions networks for the action and value heads. Table \ref{tab:mujoco} presents the results of the three algorithms tested on each MuJoCo environment. CMAPPO wins out on all three environments, followed closely by default PPO. The GRPO algorithm, which does not use a critic network, underperforms slightly in the pure reinforcement learning task, especially in the Hopper-v4 environment.

\begin{table}[H]
\centering
\caption{Results of MuJoCo environment training. Higher is better, best value in bold.}
\label{tab:mujoco}
\begin{tabular}{lrrrrr}
\hline
Model & PPO & CMAPPO & GRPO \\
Environment & Reward & Reward & Reward \\ \hline
HalfCheetah-v4 & -150.54 & \textbf{-111.10 } &  -137.18  \\
Hopper-v4 & 1185.06 & \textbf{1960.75} &  624.86 \\
Humanoid-v4 & 2897.81 & \textbf{3201.09} &  2659.32 \\
\hline
\end{tabular}
\end{table}

\section{Results}
\label{sec:results_ft}
Fine-tuning is by definition local and its performance is measurable on a case-by-case basis. To cover as many use cases as possible, we propose to examine the results through the use of two common techniques in fine-tuning: layers freezing and transfer learning.

\subsection{Fine-tuning and Frozen Layers}
In order to retain high level patterns learned during supervised training, we can freeze parts of the model to stop the loss propagation through the network. This technique is common in large language models alignment and is employed to build the results in Table \ref{tab:frozen_ft}. We fine-tune the model with no frozen layers, 25\%, 50\% and 75\% frozen layers.

\begin{table}[H]
\centering
\caption{Results of fine-tuning models on Financial, Industrials and Technology dataset compared to the original model. In rows, the model's layers are progressively frozen. In columns, each sector represents the testing set of the model. Lower is better, best value in bold.}
\label{tab:frozen_ft}
\begin{tabular}{lrrrrrrr}
\hline
Frozen \% & Model & \multicolumn{2}{c}{Financial} & \multicolumn{2}{c}{Industrials} & \multicolumn{2}{c}{Technology} \\
Metric &  & \multicolumn{1}{c}{MSE} & \multicolumn{1}{c}{MAE} & \multicolumn{1}{c}{MSE} & \multicolumn{1}{c}{MAE} & \multicolumn{1}{c}{MSE} & \multicolumn{1}{c}{MAE} \\ \hline
\multirow{3}{*}{0\%} & PPO & 0.200 & 0.271 & 0.119 & 0.116 & 0.126 & 0.119 \\
 & CMAPPO & 0.324 & 0.208 & 0.146 & 0.160 & 0.203 & 0.189 \\
 & GRPO & 0.198 & 0.109 & 0.118 & 0.113 & 0.124 & 0.115 \\
\hline
\multirow{3}{*}{25\%} & PPO & 0.199 & 0.114 & 0.120 & 0.116 & 0.125 & 0.118 \\
 & CMAPPO & 0.300 & 0.204 & 0.202 & 0.211 & 0.525 & 0.341 \\
 & GRPO & 0.198 & 0.108 & \textbf{0.118} & \textbf{0.112} & 0.124 & 0.113 \\
\hline
\multirow{3}{*}{50\%} & PPO & 0.202 & 0.113 & 0.119 & 0.117 & 0.124 & 0.117 \\
 & CMAPPO & 0.237 & 0.155 & 0.257 & 0.248 & 0.151 & 0.151 \\
 & GRPO & \textbf{0.195} & \textbf{0.108} & \textbf{0.118} & \textbf{0.112} & 0.124 & 0.113 \\
\hline
\multirow{3}{*}{75\%} & PPO & 0.200 & 0.114 & 0.119 & 0.117 & 0.124 & 0.117 \\
 & CMAPPO & 0.270 & 0.183 & 0.289 & 0.272 & 0.137 & 0.135 \\
 & GRPO & \textbf{0.195} & \textbf{0.109} & 0.118 & 0.113 & \textbf{0.123} & \textbf{0.113} \\
\hline
Original & Backbone & 0.202 & 0.111 & 0.120 & 0.115 & 0.124 & 0.116 \\
 \hline
\end{tabular}
\end{table}

GRPO performed the best overall, either improving or leaving the backbone model unchanged. Notably, freezing at least 50\% of the encoder layers gave consistently the best performance when using GRPO. PPO proposed a minor improvement in some categories, for instance in Financial at 25\%, but mostly left the model unchanged. CMAPPO performed the worst in the fine tuning, provoking large negative changes to the model even with 75\% of the encoder frozen. 
The source of the performance of GRPO in fine-tuning is the same reason it was the worst performer in the pure reinforcement learning task: the absence of a value function. While this is mostly a disadvantage learning control tasks, in the case of fine-tuning the difference of complexity between the value network and the backbone severely hinders the performance of PPO and CMAPPO. In the case of CMAPPO, the latent representation offered by the subagents are also reconciled using an action network. This design is coherent with the original implementation of CMAPPO but also adds another layer of abstraction the model needs to learn. 
A possible improvement for PPO and CMAPPO would be to run the model without propagating the loss back to the backbone to train the value network. By delaying the learning, the value network could learn a proper representation of the advantage in the task and nudge the backbone in the right direction.

\subsection{Transfer Learning}
\label{sub:transfer_learning}
Transfer learning is a machine learning technique through which a model learns general concepts applicable across multiple datasets. We experiment on transfer learning by fine-tuning and testing the model on the three datasets. 

\begin{table}[H]
\centering
\caption{Reference values before fine-tuning.}
\label{tab:transfer_base}
\begin{tabular}{lrrrrrr}
\hline
Trained on &  \multicolumn{2}{c}{Financial} & \multicolumn{2}{c}{Industrials} & \multicolumn{2}{c}{Technology} \\
Tested on &  \multicolumn{1}{c}{MSE} & \multicolumn{1}{c}{MAE} & \multicolumn{1}{c}{MSE} & \multicolumn{1}{c}{MAE} & \multicolumn{1}{c}{MSE} & \multicolumn{1}{c}{MAE}\\ \hline
Financial & \textbf{0.203} & 0.118 & 0.207 & 0.113 & 0.203 & \textbf{0.110}\\
Industrials & 0.224 & 0.227 & 0.128 & 0.121 & \textbf{0.122} & \textbf{0.114}\\
Technology & 0.256 & 0.229 & 0.132 & 0.118 & \textbf{0.131} & \textbf{0.117} \\
\end{tabular}
\end{table}

Table \ref{tab:transfer_base} presents the results of the model on the Finance, Industrials and Technology datasets before fine-tuning. Instead of training the backbone model on all three datasets and fine-tuning for one, we train the backbone on a single dataset and test the MSE/MAE on all three. The Financial appears as the most challenging dataset, performing quite worse than the baseline when tested on Industrials and Technology. The model trained on Industrials manages to nearly match the performance of the models trained on Financial and Technology. Finally, the Technology model is by far the best, outperforming Industrials even when tested on Industrials. This metric could be interpreted as the degree of high level patterns present in the dataset. These high level patterns can be applied to any similar dataset, and ultimately are more powerful predictive tools than the past history for a given example. 

\begin{table}[H]
\centering
\caption{Results of fine-tuning models on Financial, Industrials and Technology dataset compared to the original model. The model is fine-tuned and tested on the specified sector for each row. In columns, each sector represents the original training set of the model. Lower is better, best value in bold.}
\label{tab:transfer_ft}
\begin{tabular}{llrrrrrr}
\hline
Trained on $\rightarrow$ &  & \multicolumn{2}{c}{Financial} & \multicolumn{2}{c}{Industrials} & \multicolumn{2}{c}{Technology} \\
 Fine-tuned on $\downarrow$ & Method & \multicolumn{1}{c}{MSE} & \multicolumn{1}{c}{MAE} & \multicolumn{1}{c}{MSE} & \multicolumn{1}{c}{MAE} & \multicolumn{1}{c}{MSE} & \multicolumn{1}{c}{MAE}\\ \hline
\multirow{4}{*}{Financial} & PPO & 0.279 & 0.196 & 0.213 & 0.219 & 0.247 & 0.219\\
 & CMAPPO & 0.224 & 0.143 & \textbf{0.145} & 0.152 & \textbf{0.151} & 0.151 \\
 & GRPO & 0.230 & 0.156 & 0.155 & 0.167 & 0.170 & 0.165\\
 & Baseline & \textbf{0.203} & \textbf{0.118} & 0.207 & \textbf{0.113} & 0.203 & \textbf{0.110}\\
\hline
\multirow{4}{*}{Industrials} & PPO & 0.201 & 0.115 & 0.123 & 0.119 & 0.131 &  0.121 \\
 & CMAPPO & 0.204 & 0.117 & 0.127 & 0.121 & 0.137 & 0.125 \\
 & GRPO & \textbf{0.197} & \textbf{0.110} & \textbf{0.119} & \textbf{0.113} & 0.125 & \textbf{0.114} \\
 & Baseline & 0.224 & 0.227 & 0.128 & 0.121 & \textbf{0.122} & \textbf{0.114}\\
\hline
\multirow{4}{*}{Technology} & PPO & 0.198 & 0.114 & 0.125 & 0.117 & 0.127 & 0.120 \\
 & CMAPPO & 0.202 & 0.115 & 0.128 & 0.119 & 0.129 & 0.120\\
 & GRPO & \textbf{0.189} & \textbf{0.108} & \textbf{0.118} & \textbf{0.112} & \textbf{0.123} & \textbf{0.113}  \\
 & Baseline & 0.256 & 0.229 & 0.132 & 0.118 & 0.131 & 0.117 \\
\hline
\end{tabular}
\end{table}

Table \ref{tab:transfer_ft} presents the results of the model on the Finance, Industrials and Technology datasets after fine-tuning. A first observation is the improvement in performance in all nearly all models from the baseline in Table \ref{tab:transfer_base}. Some of the most substantial gains are found in the model trained on Financial, which improved its performance in MSE for both Industrials and Technology but moreover completely dominates the MAE benchmark. On the MSE front, the model trained on Industrials had the best results and beat out the best reference values for each sector.

Notable exceptions are the model trained and fine-tuned on Financial, and the model trained on Technology and fine-tuned on Industrials. In both cases, neither PPO, CMAPPO or GRPO managed to improve the performance, and testing on unseen data yielded a worse result. For the first case, the likely explanation is an overfitting to the train data: effectively, the model was trained twice on the same dataset, once with supervised learning, and again using reinforcement learning. The second case is different: the original Technology model already performed outstandingly well in Industrials, beating out even the models trained on the complete dataset. The fine-tuning failed to further improve that performance, marking the importance of establishing baselines before introducing fine-tuning to the pipeline.

The patterns noticed in Table \ref{tab:frozen_ft} largely stand, with GRPO clearly distinguishing as the better option in nearly all cases. CMAPPO performed exceptionally well on Financial, outperforming both PPO and GRPO. The superagent managed to reconcile the actions of the subagents despite the added complexity of the actor and critic network. PPO nearly always improves on the baseline and constitute a valid choice for fine-tuning. The recommended algorithm stands out as GRPO, which uses fewer computational resources and yields the best performance. Committing to the actor paradigm and removing the critic network greatly simplifies the fine-tuning architecture, allowing for direct backpropagation through the backbone without the need for intermediary networks.

These results also clearly indicate the value of transfer learning for timeseries predictor. One of the best use case for fine-tuning appears to be adapting models from their supervised training dataset to another. This is in line with the current state of fine-tuning in large language models, which often adapts model after pre-training to diverse specific tasks. This result also highlights two clear areas for improvement in timeseries predictors: firstly, large pre-trained models can be built, and later specialized to a given dataset. But the biggest challenge to generalize this method is to specify a model and a fine-tuning environment that allows for various observation space and exogenous features.

\section{Key hyperparameters}
\label{sec:tuning_ft}
PPO and its variants are known to be sensitive to hyperparameters. In order to compare each algorithm fairly, we show in this section specific and non-specific hyperparameters tuning. All models presented from this point onward use the backbone as the action network, and a value network with 2 layers and 256 hidden dimensions when relevant (PPO, CMAPPO).

\subsection{Training time}
\begin{figure}[H]
    \centering
    \includegraphics[scale=0.3]{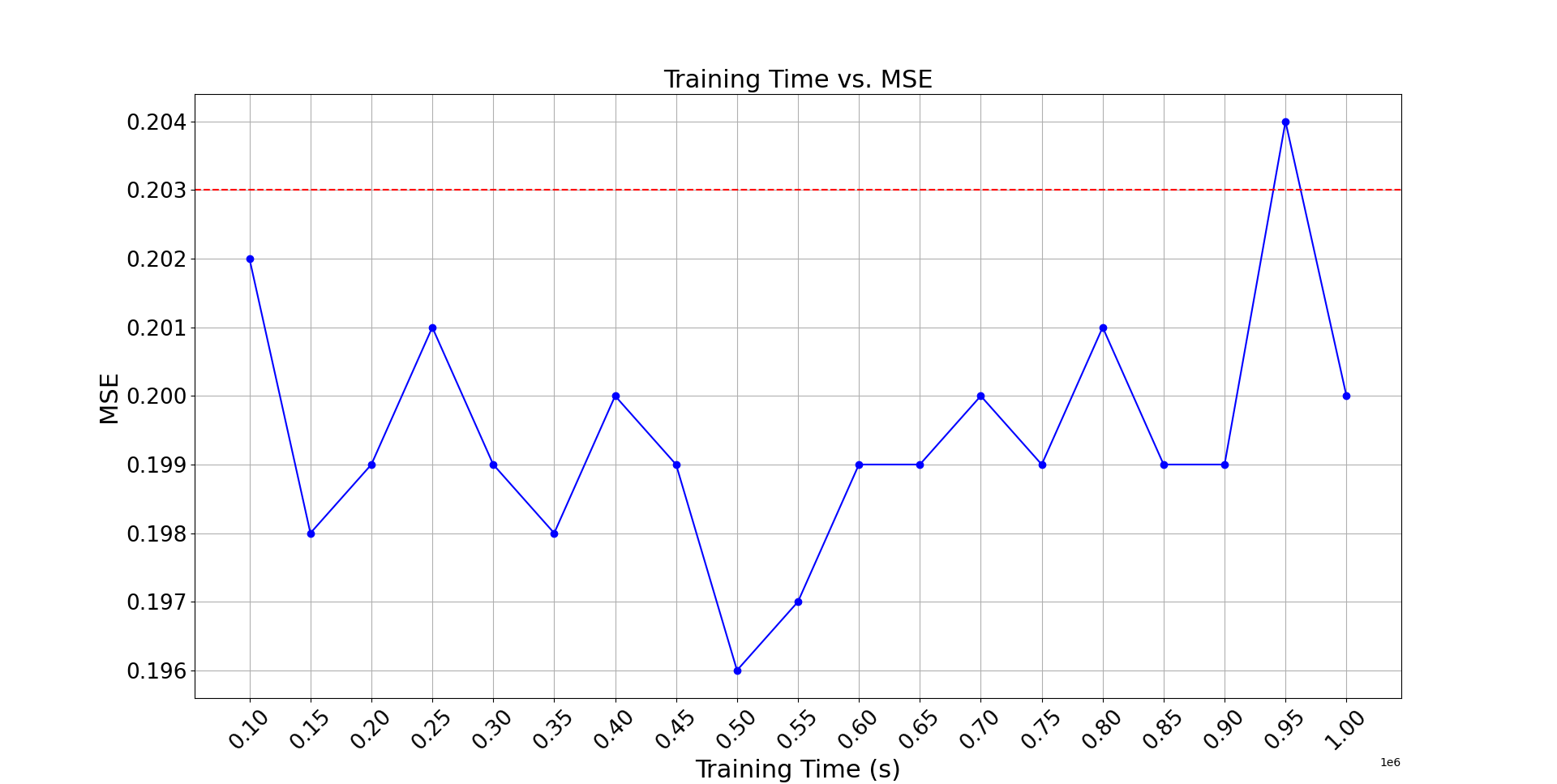}
    \caption{Training time vs MSE. Dotted line is the original model performance before fine-tuning. Training time is scaled down from 1e6 for readability.}
    \label{fig:ft_training_time}
\end{figure}

Training time is common to PPO, CMAPPO and GRPO. A higher number of timesteps will lead to a better performance in the environment until the agent reaches a plateau, at the expense of a higher computational cost. We fine-tune the model on the Financial dataset using PPO at different timesteps and plot the MSE over time in Figure \ref{fig:ft_training_time}. We found 500 000 timesteps to be the best value as a balance between overfitting and underfitting.

\subsection{Number of subagents (CMAPPO)}
\label{sub:subagents_nb_influence}
Number of subagents is specific to CMAPPO and controls how many subagents are trained before the superagent. We fine-tune a predictive model on the Financial dataset with an increasing number of subagents and test the MSE/MAE after fine-tuning. Table \ref{tab:nb_subagents} presents the MSE and MAE with increasing numbers of subagents and compared to the backbone model. We found 10 subagents to be the best configuration, despite the backbone outperforming the fine-tuned model in all configurations. 

\begin{table}[H]
\centering
\caption{The influence of the number of subagents when fine-tuning the backbone compared to the non fine-tuned backbone.}
\label{tab:nb_subagents}
\begin{tabular}{lll}
\hline
Number of Subagents &  \multicolumn{2}{c}{Financial}  \\
Metric & \multicolumn{1}{c}{MSE} & \multicolumn{1}{c}{MAE} \\ \hline
2  & 0.925 & 0.401 \\
4 & 0.461 & 0.275 \\
6 & 0.394 & 0.256 \\
8 & 0.337 & 0.211 \\
10 & \textbf{0.271} & \textbf{0.183} \\
12 & 0.283 & 0.191 \\
\hline
Backbone &  0.202 &  0.111 \\
\end{tabular}
\end{table}

\subsection{Group size (GRPO)}
Group size is specific to GRPO and determines the size of the group used to calculate the advantage. Similarly to Subesection \ref{sub:subagents_nb_influence}, we fine-tune a predictive model on the Financial dataset with an increasing group size and test the MSE/MAE after fine-tuning. Table \ref{tab:group_size} presents the results of the model at group sizes from 2 to 12. We found that a group size of 8 is optimal for both computational load and model performance. 

\begin{table}[H]
\centering
\caption{The influence of the group size when fine-tuning the model on the Financial dataset.}
\label{tab:group_size}
\begin{tabular}{lrr}
\hline
Group Size &  \multicolumn{2}{c}{Financial}  \\
Metric & \multicolumn{1}{c}{MSE} & \multicolumn{1}{c}{MAE} \\ \hline
2  & 0.200 &  0.204 \\
4 & 0.198 & 0.199 \\
6 & 0.197 &  0.199 \\
8 & \textbf{0.195} &  0.196 \\
10 & 0.199 & 0.201 \\
12 & 0.201 & 0.203 \\
\hline
Backbone &  0.202 &  \textbf{0.111} \\
\end{tabular}
\end{table}

\section{Conclusion}
\label{sec:conclusion_ft}
Fine-tuning timeseries predictors is emerging as an essential post-supervised training step to improve the performance of models. As the paradigm shifts from local models to larger, eclectic models harnessing the predictive power of many timeseries from diverse fields, fine-tuning becomes even more essential. At scale, is it far more cost effective to fine-tune a large model to a specific use case than retraining on large datasets. As the computational load for supervised learning gets higher and the models get larger, which has been the trend observed in LLMs and timeseries predictors, fine-tuning becomes even more attractive.

There are still several limitations, the most prominent being that pre-trained models use a fixed size input vector. This problem is not encountered in standard large language models, as the alphabet is tokenized to represent the entirety of the model output. But timeseries prediction is a continuous process, and further innovation is needed in foundational model to break out of fixed size vectors and scale up the models on large datasets, without relying on tricks such as projection layers. In the same spirit, architectural changes to foundational model allowing for variable output vector size would benefit the industry integration of timeseries predictors.

The environment definition and reward structure are key to the success of fine-tuning. Empirically, we noticed better results by bounding the reward to values between -1 and 1. The algorithm used is also a determining factor, and GRPO emerges as the clear winner in this chapter. This result is in line with the recent advances in LLMs, and further strengthens the conjecture that LLMs and timeseries predictors based on the same architecture share scaling features. If this conjecture reveals to be true, timeseries predictors are in a fantastic second mover position to implement even more innovations the thriving LLM community is building.
\printbibliography

\appendix

\end{document}